\documentclass[journal,twoside,web]{ieeecolor}
\usepackage{generic}
\usepackage{cite}
\usepackage{amsmath,amssymb,amsfonts}
\usepackage{algorithmic}
\usepackage{graphicx}
\usepackage{subfigure}
\usepackage{algorithm,algorithmic}
\usepackage{hyperref}
\usepackage{makecell}
\hypersetup{hidelinks=true}
\usepackage{textcomp}

\usepackage{xcolor}
\makeatletter

\newcommand{\Rmnum}[1]{\expandafter\@slowromancap\romannumeral #1@}

\newcommand{\blue}[1] {\textcolor[rgb]{0.0,0.0,0.0}{{#1}}}
\makeatother

\def\BibTeX{{\rm B\kern-.05em{\sc i\kern-.025em b}\kern-.08em
    T\kern-.1667em\lower.7ex\hbox{E}\kern-.125emX}}
\begin{document}
\title{HFS-TriNet: A Three-Branch Collaborative Feature Learning Network for Prostate Cancer Classification from TRUS Videos}
\author{Xu Lu, Qianhong Peng, Qihao Zhou, Shaopeng Liu, Xiuqin Ye, Chuan Yang, Yuan Yuan
\thanks{This work was supported in part by Key-Area Research and Development Program of Guangdong Province (2023B0303020001); The National Natural Science Foundation of China (62176067);  Scientific and Technological Planning Project of Guangzhou (202103000040, 2023B03J1378); Research project of Guangdong Polytechnic Normal University (22GPNUZDJS14); Guangdong Provincial Key Laboratory (2018B030322016); Industry-University-Research Innovation Fund for Chinese Universities (2023HT023); Guangzhou Basic and applied basic research project(2024A04J4762).(Corresponding author:Xiuqin Ye, Chuan Yang, Yuan Yuan.)}
\thanks{Xu Lu, Yuan Yuan and Shaopeng Liu are with the Guangdong Polytechnic Normal University, Guangzhou 510665, China; Guangdong Provincial Key Laboratory of Intellectual Property \& Big Data, Guangzhou 510665, China (e-mail: xulu@gpnu.edu.cn; yuanyuan@gpnu.edu.cn; liushaopeng@gpnu.edu.cn).}
\thanks{Qianhong Peng and Qihao Zhou are with the Guangdong Polytechnic Normal University, Guangzhou 510665, China (e-mail: qianhongpeng@stu.gpnu.edu.cn; zhouqihao@stu.gpnu.edu.cn).}
\thanks{Chuan Yang is with the Department of Ultrasond, The First Affiliated Hospital of Jinan University, Guangzhou 510630, China(e-mail: yc1005@jnu.edu.cn).}
\thanks{Xiuqin Ye is with the Department of Ultrasond, Shenzhen People's Hospital, Shenzhen 518020, China (e-mail: yexiuqin.us@szhospital.com)}}

\maketitle

\begin{abstract}
Transrectal ultrasound (TRUS) imaging is a cost-effective and non-invasive modality widely used in the diagnosis of prostate cancer. 
The computer-aided diagnosis (CAD) relying on TRUS images has been extensively investigated recently.
Compared to static images, TRUS video provides richer spatial-temporal information, which make it a promising alternative for improving the accuracy and robustness of CAD systems. 
However, TRUS video analysis also introduces new challenges.
These include information redundancy, which increases computational costs; high intra- and inter-class similarity, which complicates feature extraction; and a low signal-to-noise ratio, which hinders the identification of clinically relevant information.
To address these problems, we propose a heuristic frame selection (HFS) and a three-branch collaborative feature learning network (HFS-TriNet) for prostate cancer classification from TRUS videos.
Specifically, selecting a clip of video frames at intervals for training can mitigate redundancy.
The HFS strategy dynamically initializes the starting point of each training clip, which ensures that the sampled clips span the entire video sequence.
For better feature extraction, besides a regular ResNet50 branch, we also utilize 1) a large model branch based a pre-trained medical segment anything model (SAM) to extract deep features of each frame and a normalization-based attention module to explore the temporal consistency; and 2) a wavelet transform convolutional residual (WTCR) branch that extracts lesion edge information in the high-frequency domain and performs denoising in the low-frequency domain.
\blue{
Extensive experiments on a multi-institutional TRUS video dataset, using patient-level evaluation protocols, demonstrate that the proposed HFS-TriNet achieves improved and balanced performance across multiple evaluation metrics, indicating its potential utility for TRUS video--based computer-aided decision support.
}
\end{abstract}


\section{Introduction}
\IEEEPARstart{P}{rostate} cancer is one of the most prevalent malignancies among men and a leading cause of cancer-related deaths worldwide~\cite{r1,r2}. 
Early detection is critical for improving prognosis and reducing mortality. 
Among current diagnostic modalities, transrectal ultrasound (TRUS) is widely used for its advantages of being non-invasive, cost-effective, and convenient. 
These features make TRUS particularly suitable for large-scale early screening of prostate cancer~\cite{r2}. 
However, the accuracy of TRUS-based diagnosis heavily depends on the experience and skill of the physician, which can lead to variability in clinical outcomes~\cite{r3}.
This highlights the urgent need for computer-aided diagnostic (CAD) systems, particularly computer-aided decision-support tools, that can enhance the reliability and consistency of TRUS-based prostate cancer detection while keeping clinicians in the decision loop.

Recent years has been witnessed growing interest in CAD techniques based on TRUS imaging. 
Most existing methods~\cite{r29,r30,r31,r32,r33,r34,r61} rely on static ultrasound images, which represent only a limited snapshot of the prostate and may fail to capture the full diagnostic context. 
In contrast, TRUS videos contain richer spatio-temporal information and can provide a more comprehensive representation of tissue dynamics and morphology~\cite{r62,r63}. 
Moreover, video-based analysis avoids the subjective and potentially biased process of manually selecting representative frames, which often leads to information loss. 
These advantages make TRUS video a promising alternative for improving the accuracy and robustness of CAD systems~\cite{r64}.

Despite its potential, TRUS video analysis also introduces several unique challenges. 
First, consecutive video frames are highly redundant, as they often contain minimal changes in anatomical structures. 
To mitigate computational cost and over-fitting, it is necessary to sample frames at intervals to form informative clips for training. 
However, ensuring that sampled clips contain sufficient prostate-relevant content is non-trivial. 
For example, in patients with malignant non-diffuse prostate cancer, lesion-related information may only appear in a small portion of the video~\cite{r65,r66}.
Second, TRUS video data exhibit high variability and complexity due to inter-patient anatomical differences and pathological deformations. 
The overlapping nature of prostate tissue structures leads to high intra-class similarity, while early-stage prostate cancer and benign prostatic hyperplasia may share similar visual patterns, resulting in inter-class ambiguity~\cite{r67}. 
These factors significantly increase the difficulty of reliable feature extraction. 
Finally, TRUS videos are often affected by speckle noise caused by acoustic attenuation and scattering artifacts inherent to ultrasound imaging. 
In patients with diffuse prostate cancer, the widespread infiltration of lesions may blur glandular boundaries, making it difficult to accurately localize abnormal regions~\cite{r68,r69}.
These challenges hinder the performance of video-based CAD systems for prostate cancer classification~\cite{r70}. 

\blue{
To address these challenges, we propose \emph{HFS-TriNet}, a problem-driven framework for prostate cancer classification from TRUS videos, which explicitly targets the unique characteristics and clinical constraints of TRUS video analysis.
Specifically, we introduce a heuristic frame selection (HFS) strategy that mitigates temporal redundancy while promoting comprehensive temporal coverage of prostate-relevant information.
By adaptively initializing clip starting points, HFS reduces the likelihood of missing lesion-related content, particularly in non-diffuse prostate cancer cases.
}
\blue{
For robust feature representation, HFS-TriNet adopts a three-branch collaborative learning architecture.
In addition to a baseline ResNet50 branch for spatio-temporal feature extraction, we incorporate
1) a large-model branch based on a pre-trained medical Segment Anything Model (SAM)~\cite{r47} to inject high-level semantic priors, together with a normalization-based attention module that adapts frozen SAM features to TRUS-specific characteristics; and
2) a wavelet transform convolutional residual (WTCR) branch that explicitly models ultrasound frequency-domain properties, enhancing lesion boundary cues in the high-frequency domain while suppressing speckle noise through low-frequency denoising.
}
\blue{
Experimental results on a multi-institutional TRUS video dataset indicate that the proposed framework achieves stable and competitive performance under patient-level evaluation, supporting its potential utility for TRUS video–based computer-aided decision support.
}

\blue{
The main contributions of this work are summarized as follows:
}
\begin{itemize}
	\item \blue{
	We propose a heuristic frame selection (HFS) strategy tailored for TRUS videos, which ensures effective temporal coverage under fixed clip-length constraints and improves utilization of prostate-relevant information without introducing additional trainable parameters.
	}
	
	\item \blue{
	We develop a three-branch collaborative feature learning framework that integrates spatio-temporal modeling, large-model semantic priors, and frequency-domain feature enhancement, with each branch explicitly designed to address a distinct challenge inherent to TRUS video analysis.
	}
	
    \item \blue{
	We perform an extensive evaluation on a multi-institutional TRUS video dataset under patient-level protocols, including cross-validation and statistical analysis, to assess the effectiveness and stability of the proposed framework for prostate cancer classification.
}
\end{itemize}

\blue{
This work presents a clinically motivated, system-level framework for TRUS video–based prostate cancer classification.
By aligning heuristic sampling, semantic modeling, and frequency-aware representation with the characteristics of TRUS videos, HFS-TriNet provides an effective approach for video-level analysis under realistic clinical acquisition conditions.
}


\label{sec:introduction}

\section{Related Work}
We conducted a review of related literature from the following three perspectives: 1) prostate cancer analysis based on ultrasound images, 2) prostate cancer analysis based on ultrasound videos, and 3) the application of large-scale models in ultrasound image feature extraction.

\subsection{Prostate Cancer Analysis Based on Ultrasound Image}
Research on prostate cancer ultrasound image analysis has progressively shifted from conventional image processing techniques to advanced deep learning paradigms. For instance, Lu et al.~\cite{r29} developed an automatic region-based Gleason grading framework, which achieved a Kappa concordance coefficient of 0.86 with the pathological ground truth by jointly localizing cancerous regions and performing histopathological grading. Subsequently, Lu et al.~\cite{r30} proposed a multiscale generative adversarial network enhanced with depth-perceptual loss, which improved the resolution of prostate ultrasound images by a factor of four and enhanced anatomical edge sharpness by 41\%. In another study, Lu et al.~\cite{r31} introduced a self-supervised dual-head attentional framework that extracts robust spatio-temporal features from unlabeled TRUS videos via dual-branch collaborative learning, significantly reducing the reliance on manual annotations. In addition, Lu et al.~\cite{r32} proposed a deep metric learning network that enhances cross-center generalization in prostate cancer classification by jointly optimizing feature space representations and data augmentation strategies.
Although these methods have made significant progress in lesion localization, image enhancement, self-supervised learning, and cross-center generalization, they are primarily based on static images or single modalities. As a result, they lack deep modeling of temporal continuity and dynamic features across consecutive frames, limiting their applicability in ultrasound video analysis tasks.
To address common issues such as speckle noise and low contrast inherent in ultrasound imaging, Yu et al.~\cite{r33} designed a speckle-reducing anisotropic diffusion algorithm that adaptively adjusts diffusion coefficients to suppress noise in low-contrast regions while preserving edge structures. Regarding cross-device generalization, Pérez et al.~\cite{r34} proposed a deep transfer learning method combined with feature disentanglement, which alleviates performance degradation caused by variations in imaging parameters among different ultrasound systems (e.g., GE, Philips, Siemens), thereby improving generalization across platforms.

However, these methods mostly focus on image-level preprocessing or adaptation in the feature space and have yet to incorporate rich spatio-temporal contextual information into the modeling process. Consequently, they still face limitations when dealing with dynamic structural changes or temporally distributed video data.

\subsection{Prostate Cancer Analysis Based on Ultrasound Video}
Ultrasound video analysis for prostate cancer diagnosis has gained considerable attention in recent years. Compared to static images, ultrasound videos offer richer spatio-temporal information but also pose challenges such as speckle noise and anatomical variability~\cite{r35}. To overcome these limitations, numerous deep learning frameworks have been introduced. For example, Zhang et al.~\cite{r36} introduced MedMusNet, which employs mask-enhanced deep supervision to achieve accurate detection of clinically significant prostate cancer in B-mode micro-ultrasound images, reaching an accuracy of approximately 76\%. Nevertheless, the method reports a relatively low Dice similarity coefficient, indicating room for improvement in lesion localization precision, and its robustness against image noise remains limited. Wu et al.~\cite{r37} developed a multimodal classification framework that integrates B-mode and elastography imaging, substantially enhancing the identification of clinically significant prostate cancer. However, the model's capacity for modeling temporal dynamics in ultrasound videos is insufficient, and it similarly lacks validation of generalization across different scanners and clinical centers. Sun et al.~\cite{r38} utilized a 3D convolutional neural network to process transrectal ultrasound videos, achieving high diagnostic accuracy that even outperformed experienced radiologists. However, the method relies on static structural segmentation and regression, which limits its robustness to dynamic factors such as variations in heart rate, potentially affecting its stability in complex clinical scenarios. Orlando et al.~\cite{r40} proposed a deep learning-based system for automatic 3D prostate segmentation, which demonstrated good generalization across diverse clinical datasets. Nonetheless, its adaptability to ultrasound noise is limited, and segmentation accuracy tends to decrease under conditions of blurred boundaries and strong artifacts.

In summary, while these existing methods have achieved promising results in classification and segmentation tasks for prostate ultrasound images and videos, they generally suffer from inadequate modeling of dynamic information in ultrasound videos, limited robustness to noise, and insufficient generalization across imaging devices.

\subsection{Large-Scale Models for Feature Extraction in Ultrasound Imaging}
In recent years, with the rapid development of deep learning, models that fuse convolutional neural networks (CNNs) with Transformer architectures have achieved significant progress in ultrasound image analysis. These approaches aim to strike a balance between spatial structural modeling and long-range dependency learning, thereby enhancing both classification and segmentation performance. Zhao et al.~\cite{r41} proposed AResNet-ViT, which integrates CNNs and Vision Transformers through a dual-branch architecture, thereby enhancing local structural perception and global semantic representation for breast nodule classification. However, this method still relies on static image feature extraction and lacks modeling of dynamic information between consecutive frames, limiting its applicability to ultrasound video diagnosis. Mo et al.~\cite{r42} introduced HoVer-Trans, which incorporates anatomical priors and decouples horizontal and vertical features to enable ROI-free breast cancer diagnosis. Nonetheless, the method heavily depends on the structural consistency of input images, which may result in degraded feature representations under varying acquisition devices or complex tissue structures.

In terms of lightweight architecture design, Sivasubramanian et al.~\cite{r43} proposed an attention-based lightweight CNN framework for rapid fetal ultrasound plane recognition. Although the method is efficient, its shallow architecture lacks the ability to integrate multi-scale contextual information, making it less effective when handling clinically complex images with fine structural details or ambiguous boundaries. Wang et al.~\cite{r44} presented UltraSegNet, which combines multi-head attention and deformable convolutions to enhance the representation of local features for breast lesion detection. However, its modeling of lesion deformation remains limited to static feature stacking, lacking temporal modeling of lesion progression. Fazry et al.~\cite{r46} proposed UltraSwin to capture spatiotemporal dynamics from echocardiogram videos for ejection fraction estimation. Nevertheless, the model performs regression without explicit segmentation of cardiac structures, leading to weak sensitivity to anatomical changes and instability under varying heart rates or abnormal rhythms.

Huang et al.~\cite{r47} conducted a systematic evaluation of SAM in the context of medical imaging and found that its direct transfer to multi-modal medical images suffers from significant performance degradation. Although SAM demonstrates strong zero-shot segmentation capabilities, it performs inconsistently when dealing with challenges specific to ultrasound—such as fuzzy anatomical structures and weak tissue boundaries—indicating that its universal segmentation strategy may not be directly applicable to the medical domain.

Overall, existing methods primarily focus on static images and fall short in capturing the dynamic yet structurally similar patterns present in ultrasound videos. Additionally, their ability to integrate multi-dimensional features remains limited.

\vspace{0.1in}
\blue{
In summary, existing TRUS-based prostate cancer studies mainly fall into static image–based methods and generic video modeling approaches.
Image-based methods rely on selected representative frames and are limited in capturing temporal context, while most video-based methods are adapted from natural video recognition and do not explicitly address TRUS-specific challenges such as temporal redundancy, speckle noise, and semantic ambiguity.}

\blue{Moreover, prior works typically improve isolated components, such as temporal aggregation or feature extraction, without a unified design that jointly considers sampling strategy, semantic representation, and ultrasound-specific characteristics.}

\blue{Motivated by these limitations, this work adopts a problem-driven, system-level perspective by integrating heuristic frame selection, large-model semantic priors, and frequency-aware feature enhancement into a collaborative framework tailored for TRUS video–based prostate cancer classification.
}

\label{sec:related}

\section{Proposed Method}
\begin{figure*}[htbp]
	\centering
	\includegraphics[width=\textwidth]{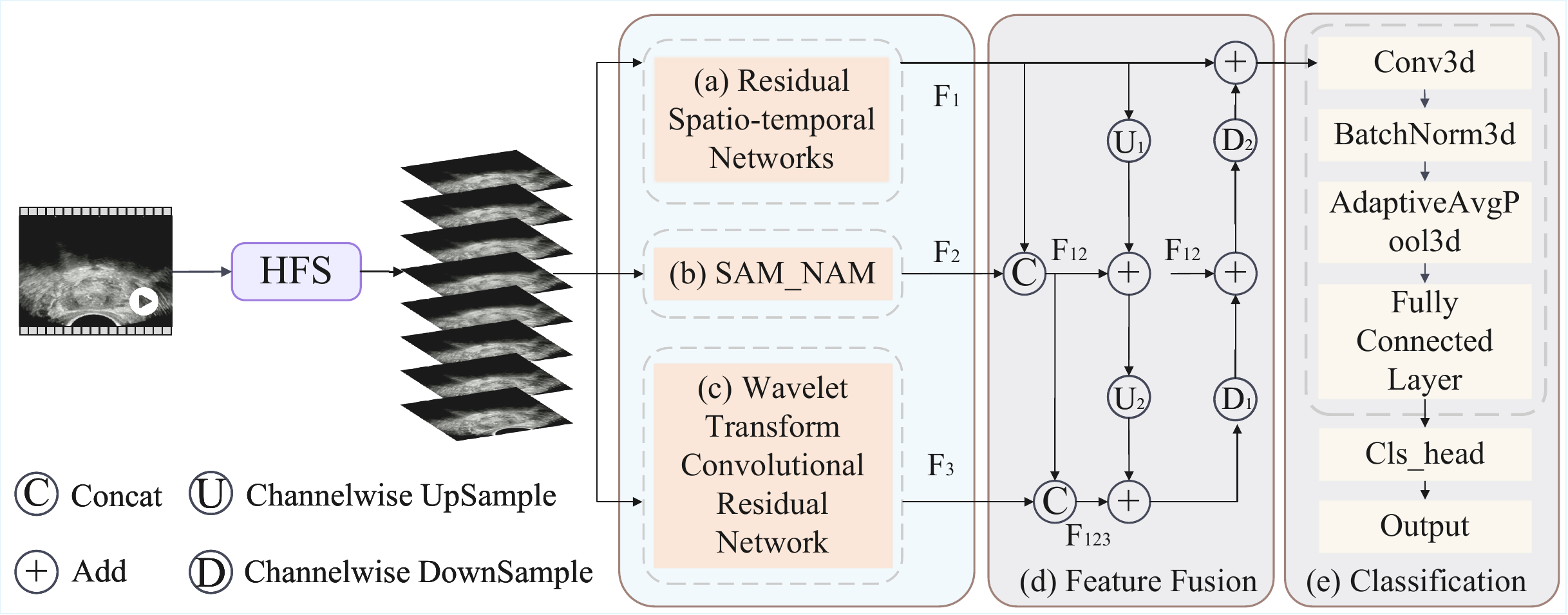}
	\caption{\blue{
	Overall architecture of the proposed HFS-TriNet.
	TRUS videos are first processed using the heuristic frame sampling (HFS) strategy to generate representative training clips.
	Each clip is then fed into three parallel feature extraction branches, whose outputs are fused using a pyramid fusion strategy, as illustrated in (d).
	The fused representation is finally passed to a classification head to produce the video-level prediction, as shown in (e).
	In the fusion diagram, $\mathrm{U}$ and $\mathrm{D}$ denote channel-wise upsampling and downsampling operations implemented via $1{\times}1{\times}1$ convolutions, respectively, without altering the spatial or temporal resolutions of the feature maps.
	}}
	\label{fig:roughly}
\end{figure*}

The overall architecture of the proposed HFS-TriNet is illustrated in Fig.~\ref{fig:roughly}.
TRUS videos are first processed using an HFS strategy to extract representative training clips with balanced temporal coverage.
Each clip is then passed through three parallel branches for feature extraction: 
(a) a backbone residual network that captures standard spatial-temporal features;
(b) a pre-trained medical SAM model for extracting deep semantic representations, followed by a normalization-based attention module to facilitate spatio-temporal fusion;
and (c) a WTCR branch, which leverages the multi-scale and multi-frequency properties of wavelet transform to extract lesion edge information and reduce speckle noise.
\blue{
Each branch is not designed as an independent enhancement, but as a complementary pathway targeting a specific limitation of TRUS video analysis.
Their coordinated integration reflects a problem-driven, system-level design that emphasizes complementary functionality across branches.
}
The features extracted from all these three branches are subsequently fused using a pyramid fusion strategy.
The fused representation is finally passed to a classification head to produce the final prediction.
We describe each component of the architecture in detail in the following sections.

\subsection{Heuristic Frame Sampling Strategy}
%
In TRUS-based learning frameworks, videos are commonly sampled at fixed intervals to generate representative clips for training.
\blue{
However, due to large variability in video length and the presence of substantial temporal redundancy, purely random or uniform sampling may inadequately capture prostate-relevant content across the entire acquisition.
}
\blue{
To accommodate these practical constraints, we design a heuristic frame sampling (HFS) strategy that determines the initial sampling positions as a function of video length, aiming to improve temporal coverage under fixed clip-length settings.
}

Let $L$ denote the total number of frames in a TRUS video, and $N$ represent the number of clips to be sampled. 
Each clip contains $T$ frames, sampled at a fixed interval of $t$ frames. 
For the $i$-th clip, where $i \in \{1, 2, \ldots, N\}$, the sampling process starts from a selected initial frame $s_i$. 
The subsequent frames within the clip are sampled as:
\begin{eqnarray}
	f_{i,j} = (s_i + j \cdot t) \bmod L, \quad j = 0, 1, \ldots, T-1,
	\label{eq:hfs}
\end{eqnarray}
where $\bmod$ denotes the modulo operation, which allows for cyclic sampling when the video length $L$ is insufficient to accommodate the full sampling range. 
\blue{This formulation enables the sampling process to wrap around to earlier frames when necessary, thereby maintaining a fixed clip length without discarding short videos.
The choice of the initial frame $s_i$ directly determines the temporal coverage and information distribution within each clip. 
To accommodate varying video lengths and improve temporal coverage of prostate-relevant content, we adopt a heuristic strategy to determine $s_i$ as a function of the video length $L$.}
The strategy is divided into the following three cases:

\paragraph{Case 1: Long videos ($L \geq N + (T - 1) \cdot t$)} 
In this case, the video is sufficiently long to allow evenly distributed and non-overlapping clips. 
We first define a sampling interval $\Delta$ for the clip starting points as:
\begin{equation}
	\Delta = \left\lfloor \frac{L - (T - 1) \cdot t}{N} \right\rfloor,
	\label{eq:Delta}
\end{equation}
where $\lfloor \cdot \rfloor$ denotes the floor (round-down) operation.

Then, the initial frame for the $i$-th clip is defined as:
\begin{equation}
	s_i = (i - 1) \cdot \Delta + U(0, \Delta),
\end{equation}
where $U(a, b)$ denotes an integer uniformly sampled from the interval $[a, b]$.
This formulation ensures that the starting frames $s_i$ are evenly spaced across the first $L - (T - 1) \cdot t$ frames, with a small amount of randomness added via $U(0, \Delta)$ to introduce variation across training epochs. 
As a result, the sampled clips can achieve full coverage of the video with minimal overlap.

\paragraph{Case 2: Medium-length videos ($ (T - 1) \cdot t \leq L < N + (T - 1) \cdot t$)} 
In this scenario, the video is not long enough to sample $N$ start points without overlap in the first $L - (T - 1) \cdot t$ frames. 
To ensure temporal coverage across the entire video and randomness, we define the starting point $s_i$ for the $i$-th clip as:
\begin{equation}
	s_i = 
	\begin{cases}
		i, & \text{if } i < L - (T - 1) \cdot t \\
		U\Big(L - (T - 1) \cdot t, L\Big), & \text{otherwise}
	\end{cases}
\end{equation}
This ensures that the earlier clips are able to guarantee full temporal coverage and non-overlapping, while the remaining ones are sampled with flexibility and variability.

\paragraph{Case 3: Short videos ($L < (T - 1) \cdot t$)} 
In this extreme case, the video is too short to contain even a single full-length clip using the regular interval $t$. 
Therefore, we allow cyclic sampling and randomly select the starting point as:
\begin{equation}
	s_i = U(1, L),
\end{equation}
ensuring that the entire video content is still utilized during training through wrap-around sampling.


\blue{
By sampling a fixed number of frames per clip, the proposed HFS strategy facilitates consistent training from variable-length TRUS videos without explicitly modeling video duration.
}
Under this formulation, the sampled clips are distributed across the entire video sequence with improved temporal coverage, which enhances the utilization of prostate-relevant content for downstream classification.
The effectiveness of the HFS strategy was empirically evaluated on annotated TRUS videos from 69 malignant non-diffuse prostate cancer patients.
Compared with random sampling, HFS reduced the proportion of clips containing only background frames from 26.1\% to 2.9\%, indicating improved sampling relevance (see the Experimental Section for details).

\begin{figure}[htbp]
	\centering
	\includegraphics[width=\linewidth]{details.pdf}
	\caption{\blue{
	Detailed architecture of the three feature extraction branches in HFS-TriNet:
	(a) a backbone residual spatio-temporal network;
	(b) a normalization-based attention module applied to features extracted from a frozen pre-trained medical SAM encoder (denoted as SAM$\_$NAM);
	and (c) a wavelet transform convolutional residual (WTCR) branch.
	}}
	\label{fig:details}
\end{figure}

\subsection{Three Collaborative Feature Extraction Branches}
We next design a three-branch collaborative architecture to extract complementary features from TRUS video clips. 
The three branches include: 1) a conventional ResNet50-based backbone branch for general spatio-temporal feature extraction, 2) a SAM$\_$NAM branch that leverages a pre-trained large medical SAM model (MedSAM) and a normalization-based attention module for enhanced semantic and temporal feature representation, and 3) a WTCR branch that utilizes the multi-scale and frequency-aware properties of wavelet transforms to capture fine-grained lesion boundaries and suppress speckle noise.
In the following, we describe each of these branches in detail.

\subsubsection{Residual Spatio-Temporal Networks}
The residual spatio-temporal network serves as the backbone of our feature extraction module, aiming to capture both intra-frame spatial semantics and inter-frame temporal dynamics from TRUS video clips. 
As shown in Figure~\ref{fig:roughly}(a), the network adopts a hybrid design that combines 2D and 3D operations~\cite{r49}.

In the spatial extraction stage, each video frame is first processed by a series of 2D ResNet blocks~\cite{r71}. 
This stage encodes global spatial context while reducing redundancy in spatial resolution.
To model motion information across frames, the spatial features are subsequently passed through grouped 3D convolutions and 3D max pooling layers.
The grouped structure allows the network to efficiently learn fine-grained temporal correlations while limiting computational cost. 
Overall, this hierarchical spatio-temporal encoding enhances the model's ability to represent both appearance and motion, and provides complementary features to those learned by the other branches.

\subsubsection{SAM\_NAM Branch}
MedSAM is a medical adaptation of the SAM~\cite{r47}, tailored for the unique characteristics of medical images, such as low contrast, anatomical variability, and limited labelled data.
The encoder is based on a Vision Transformer (ViT)~\cite{r72} backbone, which tokenizes input images into patches and applies self-attention mechanisms to model global spatial dependencies. 
Unlike traditional CNNs, ViT-based architectures can capture long-range contextual information, which is especially beneficial for recognizing diffuse or subtle lesion patterns in prostate ultrasound.

In our design, as shown in Fig.~\ref{fig:details}(b), the MedSAM image encoder processes each TRUS frame independently and outputs high-dimensional semantic feature maps that encode rich anatomical and contextual cues. 
During training, the encoder weights are frozen to preserve the strong medical priors learned from large-scale medical segmentation tasks. 
This not only improves generalization but also reduces computational overhead.

To adapt the high-dimensional semantic features extracted by MedSAM to our task-specific fusion architecture, we apply a two-stage dimensionality reduction and refinement process using normalization-based attention modules (NAM).
Let $F_{\text{SAM}}$ denote the feature maps extracted from the frozen MedSAM encoder. 
First, a batch normalization (BN) layer is applied to capture global spatio-temporal dependencies across video frames. 
The BN output is passed through a sigmoid activation to generate an attention mask, which is then element-wise multiplied with the original feature to produce the BN-enhanced feature map:
\begin{equation}
	F_{\text{BN}} = \sigma\big( \text{BN}(F_{\text{SAM}}) \big) \cdot F_{\text{SAM}},
\end{equation}
where $\sigma(\cdot)$ denotes the sigmoid function.  
This operation enables temporal context fusion by aggregating statistical variations across the batch dimension, thus enhancing global feature representation.

Next, we perform pixel-wise normalization on $F_{\text{BN}}$ to refine local spatial details. 
The normalized feature at each spatial location $(i,j)$ is defined as:
\begin{equation}
	F_{\text{norm}}(i,j) = \frac{F_{\text{BN}}(i,j)}{\sqrt{\frac{1}{C} \sum_{c=1}^{C} F_{\text{BN}}^c(i,j)^2 + \epsilon}},
\end{equation}
where $C$ is the number of channels and $\epsilon$ is a small constant for numerical stability. 
A sigmoid activation is then applied to $F_{\text{norm}}$ to generate a pixel-level attention map, and the final output of the SAM\_NAM branch is computed as:
\begin{equation}
	F_2 = \sigma(F_{\text{norm}}) \cdot F_{\text{BN}}.
\end{equation}
The pixel-level normalization-based attention module enhances the model’s sensitivity to fine-grained spatial variations and subtle lesion features.

Overall, NAM adaptively emphasizes informative regions while maintaining low computational and parameter overhead. 
Compared to conventional attention mechanisms, NAM adopts a simplified formulation based on normalization operations, making it significantly more lightweight and efficient.	
By applying NAM, the MedSAM-derived features are better aligned with the temporal and frequency-aware representations produced by the other two branches, facilitating effective multi-branch fusion.

\subsubsection{Wavelet Transform Convolutional Residual Branch}
	As shown in Fig.~\ref{fig:details}(c), we propose a frequency-aware feature extraction branch based on cascaded wavelet transform (WT) convolutions~\cite{r50}. This design aims to improve consistency in noisy and low-SNR TRUS videos by leveraging the multi-scale and directional sensitivity of wavelet decompositions.

To implement this, we first apply the 2D Haar wavelet transform to each input frame tensor $X \in \mathbb{R}^{C \times H \times W}$ using four separable filters:
\begin{align}
	f_\text{LL} &= \frac{1}{2} \begin{pmatrix} 1 & 1 \\ 1 & 1 \end{pmatrix}, \quad
	f_\text{LH} = \frac{1}{2} \begin{pmatrix} 1 & -1 \\ 1 & -1 \end{pmatrix}, \\ \nonumber
	f_\text{HL} &= \frac{1}{2} \begin{pmatrix} 1 & 1 \\ -1 & -1 \end{pmatrix}, \quad
	f_\text{HH} = \frac{1}{2} \begin{pmatrix} 1 & -1 \\ -1 & 1 \end{pmatrix}.
\end{align}
These filters are applied via depth-wise convolution to decompose the input into four frequency components:
\begin{align}
	&[X_\text{LL}, X_\text{LH}, X_\text{HL}, X_\text{HH}] = \text{WT}\left( X \right) \\ \nonumber
	&= \text{Conv2D}\big(\left[f_\text{LL}, f_\text{LH}, f_\text{HL}, f_\text{HH} \right], X \big),
\end{align}
where $X_\text{LL}$ denotes the low-frequency approximation, and $X_\text{LH}, X_\text{HL}, X_\text{HH}$ capture the horizontal, vertical, and diagonal high-frequency details, respectively. 
Each component has a spatial resolution that is half the size of $X$.

To capture frequency information at multiple scales, we recursively apply the WT to the low-frequency subband:
\begin{equation}
	[X^{(i)}_\text{LL}, X^{(i)}_\text{LH}, X^{(i)}_\text{HL}, X^{(i)}_\text{HH}] = \text{WT}(X^{(i-1)}_\text{LL}), \quad i = 1, 2, \dots, M,
\end{equation}
where $X^{(0)}_\text{LL} = X$ and $M$ denotes the number of decomposition levels.
This progressive decomposition increases frequency resolution while reducing spatial resolution, enabling hierarchical representation of structural details.

At each level, the high-frequency subbands $X^{(i)}_\text{HF} = [X^{(i)}_\text{LH}, X^{(i)}_\text{HL}, X^{(i)}_\text{HH}]$ are processed using lightweight 2D convolutions to extract localized discriminative features:
\begin{equation}
	F^{(i)}_\text{HF} = \text{Conv2D}(X^{(i)}_\text{HF}).
\end{equation}
These features are aggregated with their corresponding low-frequency components via element-wise summation, enhancing the spatial coherence of the representation while preserving important frequency-domain cues.

The fused features are then passed through inverse wavelet transforms (IWT) to progressively restore the spatial resolution. The IWT is implemented as a transposed convolution using the same set of Haar filters:
\begin{equation}
	\hat{X}^{(i-1)}_\text{LL} = \text{IWT}(\hat{X}^{(i)}_\text{LL} + F^{(i)}_\text{HF}),
\end{equation}
where the initial reconstruction input is set as $\hat{X}^{(M)}_\text{LL} = X^{(M)}_\text{LL}$.

The final reconstructed representation $\hat{X}^{(0)}_\text{LL}$ for different frames are further processed using a residual spatio-temporal network (RSTN), which consists of 3D convolutional blocks designed to capture motion patterns and contextual dependencies across frames:
\begin{equation}
	F_3 = \text{RSTN}(\hat{X}^{(0)}_\text{LL}).
\end{equation}

This wavelet-based pathway complements the other two branches by incorporating frequency-domain cues into the final representation. 
It is particularly effective for distinguishing subtle prostate lesions from noisy backgrounds, enhancing the overall robustness and discriminative power of the network.

\subsection{Feature Fusion and Classification}
For the feature embeddings $F_1$, $F_2$, and $F_3$ extracted from above three branches, we then adopt a top-down pyramid fusion module~\cite{r49} to integrate multi-branch features and apply them for video classification task, as illustrated in Fig.~\ref{fig:roughly}(d-e).
\blue{Notably, the outputs of all three branches are aligned to the same spatio-temporal resolution and channel dimensionality before fusion, ensuring compatibility for subsequent multi-level aggregation.}

Specifically, features $F_1$ and $F_2$ are first concatenated along the channel dimension (denoted as $F_{12}$) and channel-wisely added to a projected version of $F_1$ via element-wise addition, yielding an intermediate fused representation.
Followed by a channel-wise upsampling operation, this intermediate feature is further aggregated with the concatenation of all three branches ($F_{123}$) to facilitate information interaction across multiple semantic levels.
To reinforce deep semantic alignment, the aggregated features are passed through a series of channel-wise downsampling operations. 
After each downsampling step, the result is sequentially added to $F_{12}$ and $F_1$ in a bottom-up manner.

\blue{Here, both channel-wise upsampling and downsampling are implemented via $1\times1\times1$ convolutional projections operating exclusively along the channel dimension, which adjust channel dimensionality while preserving the spatial and temporal resolutions of the feature maps, as illustrated by the $\mathbf{U}$ and $\mathbf{D}$ operators in Fig.~\ref{fig:roughly}(d).
This design allows semantic re-weighting and hierarchical alignment across branches without introducing spatial or temporal rescaling.}

The fused features are subsequently passed through a 3D convolutional classification head, followed by batch normalization and adaptive spatio-temporal pooling to produce a fixed-length video representation.
Finally, a fully connected layer maps the aggregated features to the classification space, yielding the final video-level prediction.

Overall, the hierarchical fusion module supports structured interaction between lower-level detailed features and higher-level semantic representations within the proposed multi-branch framework.
\label{sec:method}

\section{Experiments}
We run experiments on a multi-institution TRUS video dataset to show the effectiveness of our proposed HFS-TriNet model.	
In the following, we first introduce the dataset and experimental setup. Then, we conduct a comprehensive comparison with state-of-the-art (SOTA) methods qualitatively and quantitatively.
Finally, we perform a detailed ablation study to analyse the effectiveness of the proposed modules.

\subsection{Dataset}
The dataset used in this study was retrospectively collected from 622 patients who visited Shenzhen People's Hospital and The First Affiliated Hospital of Jinan University between 2019 and 2022, with each patient contributing between 1 and 3 TRUS videos, resulting in a total of 1083 videos\footnote{
\blue{Due to patient privacy considerations and institutional review board (IRB) regulations, the TRUS video dataset used in this study cannot be publicly released.}
}.
Each video was labeled as benign or malignant based on biopsy-confirmed histopathological results.
\blue{The dataset comprises 597 videos from 348 patients diagnosed with benign prostatic conditions and 486 videos from 274 patients diagnosed with malignant prostate cancer.
TRUS video lengths vary substantially across patients, ranging from approximately 40 to over 200 frames}, primarily due to differences in scanning duration and probe motion during acquisition.
\blue{
To reflect routine clinical acquisition conditions, no videos were excluded based on subjective image quality assessment.
During preprocessing, peripheral regions with limited diagnostic relevance were removed, retaining the central fan-shaped imaging area with a spatial resolution of $224 \times 224$ pixels.
}

\blue{
For model development and evaluation, a strict patient-level partitioning strategy was adopted.
All videos from the same patient were exclusively assigned to a single subset to prevent data leakage.
The dataset was divided into training, validation, and testing sets with an approximate patient-level ratio of 6:2:2.
The detailed distribution of patients and videos in each subset is summarized in Table~\ref{tab:dataset_split}.
}
\blue{
During training, a balanced sampling strategy with a 1:1 positive-to-negative video ratio was employed to mitigate class imbalance, while validation and testing were performed on naturally imbalanced data.
This fixed patient-level split was used as the main evaluation setting for reporting performance metrics and for computing bootstrap-based confidence intervals.
}
\blue{
In addition, a five-fold cross-validation strategy was conducted at the patient level.
Patients were evenly divided into five mutually exclusive folds, and each fold was used once for evaluation while the remaining folds were used for training.
Performance metrics were averaged across all folds.
This was conducted as a complementary analysis to assess performance stability under different patient-level partitions.
}

\renewcommand{\arraystretch}{1.2}
\begin{table}[htbp]
	\centering
	\caption{\blue{Distribution of TRUS videos and patients across the training, validation, and testing sets}}
	\label{tab:dataset_split}
	\setlength{\arrayrulewidth}{0.8pt}
	\begin{tabular}{lcccc}
		\Xhline{1.1pt}
		\textbf{Category} & \textbf{Training set} & \textbf{Validation set} & \textbf{Testing set} & \textbf{Total} \\
		\Xhline{1.1pt}
		Negative & 340 & 137 & 120 & 597 \\
		Positive & 340 & 79 & 67 & 486 \\
		\Xhline{1.1pt}
		Total videos & 680 & 216 & 187 & 1083 \\
		Number of patients & 387 & 131 & 104 & 622 \\
		\Xhline{1.1pt}
	\end{tabular}
\end{table}


\subsection{Experimental Settings}
All experiments were implemented using PyTorch 2.0.0 and conducted on a single NVIDIA GeForce RTX 4090 GPU.
To balance temporal coverage and computational cost, the video clip length and frame sampling interval were set to $T=8$ and $t=8$, respectively.
\blue{
During training, standard data augmentation strategies, including random horizontal flipping and mild geometric perturbations, were applied to increase data diversity while preserving the anatomical structure of TRUS videos.
}
The model was optimized using the AdamW optimizer with an initial learning rate of 0.01 and a weight decay of 0.01.
To stabilize training, gradient clipping with a maximum norm of 40 was applied.
A learning rate warm-up strategy was employed at the beginning of training, followed by a multi-step decay schedule to facilitate stable convergence.
The model was trained for a total of 150 epochs.
\blue{
To mitigate class imbalance during training, focal loss was adopted.
Model selection was performed based on validation performance, and the checkpoint achieving the best validation result was used for final evaluation.
}

\subsection{Comparison with SOTA Methods}

\renewcommand{\arraystretch}{1.6}
\begin{table*}[ht]
	\centering
	\caption{ Performance comparison of different methods with 95\% confidence intervals estimated by bootstrap resampling}
	\label{tab:SOTA}
	\setlength{\tabcolsep}{3.5pt}
	\setlength{\arrayrulewidth}{0.8pt}
	\begin{tabular}{l|ccccccc}
		\Xhline{1.1pt}
		\textbf{Methods} & \textbf{Acc↑} & \textbf{F1-score↑} & \textbf{AUC↑} & \textbf{Precision↑} & \textbf{Sensitivity↑} & \textbf{Specificity↑} &\textbf{P-Value}   \\
		\Xhline{0.9pt}
		TSN~\cite{r60} & 0.6323 ± 0.0233 & 0.0130 ± 0.0127 & 0.2698 ± 0.0263 & 0.2472 ± 0.2543 & 0.0067 ± 0.0066 & \underline{0.9882 ± 0.0067} & $<$0.001   \\
		I3D~\cite{r52} & \underline{0.7091 ± 0.0219} & 0.5430 ± 0.0378 & \underline{0.6755 ± 0.0276} & 0.6300 ± 0.0446 & 0.4785 ± 0.0416 & 0.8402 ± 0.0219 & $<$0.001   \\
		TRN~\cite{r59} & 0.6374 ± 0.0232 & 0.0000 ± 0.0000 & 0.4299 ± 0.0312 & 0.0000 ± 0.0000 & 0.0000 ± 0.0000 & \pmb{1.0000 ± 0.0000} & $<$0.001  \\
		SlowFast~\cite{r58} & 0.6295 ± 0.0234 & 0.1465 ± 0.0349 & 0.4366 ± 0.0288 & 0.4446 ± 0.0877 & 0.0882 ± 0.0227 & 0.9374 ± 0.0142 & $<$0.001   \\
		SlowOnly~\cite{r58} & 0.6495 ± 0.0233 & 0.2368 ± 0.0409 & 0.4822 ± 0.0310 & 0.5618 ± 0.0751 & 0.1507 ± 0.0293 & 0.9333 ± 0.0144 & $<$0.001 \\
		TPN~\cite{r49} & 0.6517 ± 0.0236 & 0.1336 ± 0.0349 & 0.5311 ± 0.0292 & \underline{0.6826 ± 0.1184} & 0.0745 ± 0.0208 & 0.9802 ± 0.0088 & $<$0.001  \\
		TANet~\cite{r56} & 0.6445 ± 0.0232 & 0.0874 ± 0.0306 & 0.5121 ± 0.0278 & 0.6308 ± 0.1508 & 0.0473 ± 0.0173 & 0.9843 ± 0.0076 & $<$0.001  \\
		Swim~\cite{r51} & 0.6398 ± 0.0231 & 0.0133 ± 0.0130 & 0.1529 ± 0.0185 & 0.6320 ± 0.4825 & 0.0067 ± 0.0066 & \pmb{1.0000 ± 0.0000} & $<$0.001   \\
		MViT~\cite{r54} & 0.6374 ± 0.0232 & 0.0000 ± 0.0000 & 0.5816 ± 0.0290 & 0.0000 ± 0.0000 & 0.0000 ± 0.0000 & \pmb{1.0000 ± 0.0000} & $<$0.001  \\
		Text4Vis~\cite{r53} & 0.6960 ± 0.0239 & \underline{0.6166 ± 0.0292} & 0.4388 ± 0.0309 & 0.5019 ± 0.0321 & \pmb{0.7410 ± 0.0340} & 0.5478 ± 0.0307 & $<$0.001  \\
		MANet~\cite{r55} & 0.4195 ± 0.0235 & 0.0555 ± 0.0199 & 0.3406 ± 0.0261 & 0.0679 ± 0.0245 & 0.0473 ± 0.0173 & 0.6313 ± 0.0294 & $<$0.001  \\
		Mult~\cite{r37} & 0.4645 ± 0.0245 & 0.3324 ± 0.0346 & 0.3812 ± 0.0278 & 0.3037 ± 0.0353 & 0.3687 ± 0.0405 & 0.5191 ± 0.0309 & $<$0.001   \\
		\Xhline{0.9pt}
		{HFS-TriNet} & \pmb{0.8632 ± 0.0171} & \pmb{0.7782 ± 0.0296} & \pmb{0.8230 ± 0.0256} & \pmb{0.9413 ± 0.0232} & \underline{0.6642 ± 0.0395} & 0.9764 ± 0.0095 &  \\
		\Xhline{1.1pt}
	\end{tabular}
\end{table*}

\renewcommand{\arraystretch}{1.6}
\begin{table*}[ht]
	\centering
	\caption{
		Performance comparison of different methods under patient-level 5-fold cross-validation, reported as mean $\pm$ standard deviation across folds.
	}
	\label{tab:5-fold}
	\setlength{\tabcolsep}{3.5pt}
	\setlength{\arrayrulewidth}{0.8pt}
	\begin{tabular}{l|cccccc}
		\Xhline{1.1pt}
		\textbf{Methods} & \textbf{Acc↑} & \textbf{Fi\_score↑} & \textbf{AUC↑} & \textbf{Precision↑} & \textbf{Sensitivity↑} & \textbf{Specificity↑} \\
		\Xhline{0.9pt}
		TSN~\cite{r60} & $0.6318 \pm 0.0403$ & $0.0259 \pm 0.0186$ & $0.2707 \pm 0.0365$ & $0.2623 \pm 0.0947$ & $0.0198 \pm 0.0153$ & $0.9811 \pm 0.0150$ \\
		I3D~\cite{r52}  & \underline{$0.7077 \pm 0.0418$} & $0.5425 \pm 0.0644$ & \underline{$0.6742 \pm 0.0344$} & $0.6242 \pm 0.0830$ & $0.4798 \pm 0.0824$ & $0.8377 \pm 0.0340$ \\
		TRN~\cite{r59}  & $0.6369 \pm 0.0400$ & $0.0225 \pm 0.0182$ & $0.4304 \pm 0.0345$ & \underline{$0.9604 \pm 0.0173$} & $0.0186 \pm 0.0151$ & $0.9427 \pm 0.0413$ \\
		SlowFast~\cite{r58}  & $0.6299 \pm 0.0401$ & $0.1646 \pm 0.0614$ & $0.4387 \pm 0.0358$ & $0.4503 \pm 0.0896$ & $0.1037 \pm 0.0462$ & $0.9327 \pm 0.0240$ \\
		SlowOnly~\cite{r58}  & $0.6493 \pm 0.0395$ & $0.2471 \pm 0.0724$ & $0.4807 \pm 0.0363$ & $0.5584 \pm 0.0983$ & $0.1632 \pm 0.0580$ & $0.9299 \pm 0.0254$ \\
		TPN~\cite{r49}  & $0.6518 \pm 0.0392$ & $0.1508 \pm 0.0580$ & $0.5310 \pm 0.0355$ & $0.6695 \pm 0.1127$ & $0.0906 \pm 0.0449$ & $0.9743 \pm 0.0142$ \\
		TANet~\cite{r56}  & $0.6449 \pm 0.0401$ & $0.1077 \pm 0.0540$ & $0.5127 \pm 0.0365$ & $0.6239 \pm 0.1158$ & $0.0656 \pm 0.0190$ & $0.9772 \pm 0.0146$ \\
		Swim~\cite{r51}  & $0.6397 \pm 0.0399$ & $0.0306 \pm 0.0224$ & $0.1547 \pm 0.0293$ & \pmb{$0.9752 \pm 0.0180$} & $0.0196 \pm 0.0153$ & \pmb{$0.9929 \pm 0.0071$} \\
		MViT~\cite{r54}  & $0.6369 \pm 0.0404$ & $0.0227 \pm 0.0183$ & $0.5788 \pm 0.0354$ & $0.0204 \pm 0.0165$ & $0.0185 \pm 0.0150$ & \underline{$0.9928 \pm 0.0072$} \\
		Text4Vis~\cite{r53}  & $0.6401 \pm 0.0399$ & \underline{$0.6149 \pm 0.0493$} & $0.4093 \pm 0.0245$ & $0.5007 \pm 0.1047$ & \pmb{$0.7903 \pm 0.0668$} & $0.5480 \pm 0.0444$ \\
		MANet~\cite{r55}  & $0.4352 \pm 0.0266$ & $0.0629 \pm 0.0249$ & $0.3402 \pm 0.0361$ & $0.0794 \pm 0.0396$ & $0.0658 \pm 0.0190$ & $0.6292 \pm 0.0243$ \\
		Mult~\cite{r37}  & $0.4651 \pm 0.0209$ & $0.3374 \pm 0.0280$ & $0.3823 \pm 0.0347$ & $0.3109 \pm 0.0383$ & $0.3741 \pm 0.0411$ & $0.5173 \pm 0.0225$ \\
		\Xhline{0.9pt}
		HFS-TriNet   & \pmb{$0.8622 \pm 0.0289$} & \pmb{$0.7766 \pm 0.0467$} & \pmb{$0.8226 \pm 0.0297$} & $0.9379 \pm 0.0288$ & \underline{$0.6592 \pm 0.0714$} & $0.9720 \pm 0.0158$ \\
		\Xhline{1.1pt}
	\end{tabular}
\end{table*}

To assess the effectiveness of the proposed method, we conducted a comprehensive comparison between HFS-TriNet and twelve state-of-the-art (SOTA) methods commonly used for video classification.
These methods include representative architectures for temporal modeling, spatio-temporal convolution, and multimodal learning, as listed in Table~\ref{tab:SOTA}.
\blue{
For fair comparison, all baseline methods were adapted to the TRUS video classification task under a unified experimental protocol.
Specifically, the same frame sampling strategy, input resolution, clip length, and training settings were applied whenever applicable, with only task-specific modifications made to the final classification heads.
}
Six evaluation metrics, including accuracy (Acc), F1-score, AUC, precision, sensitivity, and specificity~\cite{r74,r75,r76,r77,r78}, were used to provide a comprehensive assessment of classification performance.
In the context of prostate cancer analysis, precision and sensitivity reflect complementary aspects of model behavior, while F1-score and AUC capture the overall trade-off between false positives and false negatives under different decision thresholds.

\blue{Table~\ref{tab:SOTA} reports the performance comparison between HFS-TriNet and twelve state-of-the-art methods under a fixed patient-level test split.
For each method, results are presented as the mean with corresponding 95\% confidence intervals, estimated via bootstrap resampling with $n=1000$ iterations.
Overall, HFS-TriNet achieves the highest mean accuracy, F1-score, AUC, and precision among all compared methods.
Among the baseline methods, I3D~\cite{r52} achieves competitive accuracy and AUC by modeling spatio-temporal dependencies using 3D convolutions, while Text4Vis~\cite{r53} attains a strong F1-score through semantic alignment between visual features and label representations.
However, these approaches primarily emphasize either spatio-temporal modeling or semantic priors in isolation.
In contrast, HFS-TriNet integrates complementary spatio-temporal, semantic, and frequency-domain representations, resulting in a more balanced performance profile.
Notably, while several baseline methods exhibit near-perfect precision or specificity at the expense of extremely low sensitivity, HFS-TriNet maintains high precision and specificity while achieving competitive sensitivity, which is desirable for TRUS video-based prostate cancer classification under realistic class distributions.}

\blue{To assess the statistical significance of the observed improvements, paired statistical significance tests were conducted based on AUC values.
Specifically, a paired $t$-test between HFS-TriNet and the strongest baseline method, I3D, yields a statistically significant improvement ($t = 13.26$, $p < 0.001$).
Consistent results were observed when comparing HFS-TriNet with other baseline methods under identical test conditions, with all tested differences reaching statistical significance ($p < 0.001$).
}

\blue{To further assess the robustness of the proposed method under different patient-level data partitions, we additionally report patient-level five-fold cross-validation results in Table~\ref{tab:5-fold}, where all evaluation metrics are summarized as mean $\pm$ standard deviation across folds.
As shown in Table~\ref{tab:5-fold}, HFS-TriNet consistently achieves strong performance across all folds, with relatively stable accuracy, AUC, and F1-score.
These results indicate that the proposed method exhibits limited sensitivity to specific patient splits within the available multi-institutional dataset.
This cross-validation analysis complements the bootstrap-based confidence interval evaluation conducted under the fixed train/validation/test split.
Together, these two evaluation protocols provide consistent evidence supporting the statistical reliability and stability of the reported performance.
Notably, the relative performance ranking among different methods remains largely consistent between the fixed split and cross-validation settings, further suggesting that the observed performance gains of HFS-TriNet are not driven by a particular data partition.}

\subsection{Ablation Study}
We further conducted ablation studies from three perspectives: 1) to evaluate the effectiveness of each proposed module; 2) to assess the efficiency of the HFS strategy in extracting informative frames from TRUS videos; and 3) to investigate the impact of integrating features from the large-scale MedSAM model at different stages of the network.

\subsubsection{Effectiveness of the Proposed Modules}

\begin{table*}[htbp]
	\centering
	\caption{Ablation study of the proposed components and their combinations, reported as mean $\pm$ standard deviation estimated via bootstrap resampling.}
	\label{tab:modules}
	\setlength{\tabcolsep}{3pt}
	\setlength{\arrayrulewidth}{0.8pt}
	\begin{tabular}{c|ccc|cccccc}
		\Xhline{1.1pt}
		\textbf{Methods} & \textbf{HFS} & \textbf{SAM\_NAM} & \textbf{WTCR} & \textbf{Acc↑} & \textbf{F1-score↑} & \textbf{AUC↑} & \textbf{Precision↑} & \textbf{Sensitivity↑} & \textbf{Specificity↑} \\
		\Xhline{0.9pt}
		${M_{0}}$ & & & & $0.6517 \pm 0.0236$ & $0.1336 \pm 0.0349$ & $0.5311 \pm 0.0292$ & $0.6826 \pm 0.1184$ & $0.0745 \pm0.0208$ & \underline{$0.9802 \pm 0.0088$} \\
		${M_{1}}$ & \checkmark & & & $0.6896 \pm 0.0223$  & $0.2762 \pm 0.0442$  & $0.5673 \pm 0.0311$  & \underline{$0.8909 \pm 0.0590$}  & \underline{$0.1643 \pm 0.0304$} & \pmb{$0.9885 \pm 0.0065$}  \\
		${M_{2}}$ & & \checkmark & & $0.7288 \pm 0.0214$ & $0.4809 \pm 0.0426$ & $0.6481 \pm 0.0291$  & $0.7836 \pm 0.0507$ & $0.3481 \pm 0.0395$  & $0.9453 \pm 0.0140$ \\
		${M_{3}}$ & & & \checkmark & $0.6969 \pm 0.0216$ & $0.4279 \pm0.0425$ & $0.6398 \pm 0.0290$ & $0.6764 \pm 0.0565$ & $0.3141 \pm 0.0384$ & $0.9145 \pm 0.0173$ \\
		${M_{4}}$ & \checkmark & \checkmark & & \underline{$0.8511 \pm 0.0175$} & \underline{$0.7593 \pm 0.0306$} & \underline{$0.8668 \pm 0.0207$} & $0.9139 \pm  0.0279$ & $0.6504 \pm 0.0396$ & $0.9652 \pm 0.0113$ \\
		${M_{5}}$ & \checkmark & & \checkmark & $0.6865 \pm 0.0230$ & $0.6692 \pm 0.0275$ & $0.8625 \pm 0.0204$ & $0.5419 \pm 0.0317$ & \pmb{$0.8768 \pm 0.0276$} & $0.5782 \pm 0.0306$ \\
		${M_{6}}$ & & \checkmark & \checkmark & $0.7612 \pm 0.0211$ & $0.6292 \pm 0.0362$ & $0.7180 \pm 0.0270$ & $0.7185 \pm 0.0414$ & $0.5610 \pm 0.0422$ & $0.8750 \pm 0.0200$ \\
		${M_{7}}$ & \checkmark & \checkmark & \checkmark & \pmb{$0.8622 \pm 0.0289$} & \pmb{$0.7766 \pm 0.0467$} & \pmb{$0.8226 \pm 0.0297$} & \pmb{$0.9379 \pm 0.0288$} & \underline{$0.6592 \pm 0.0714$} & {$0.9720 \pm 0.0158$} \\
		\Xhline{1.1pt}
	\end{tabular}
\end{table*}

\begin{figure}[t]
	\centering
	\subfigure[]{
		\includegraphics[width=0.47\linewidth]{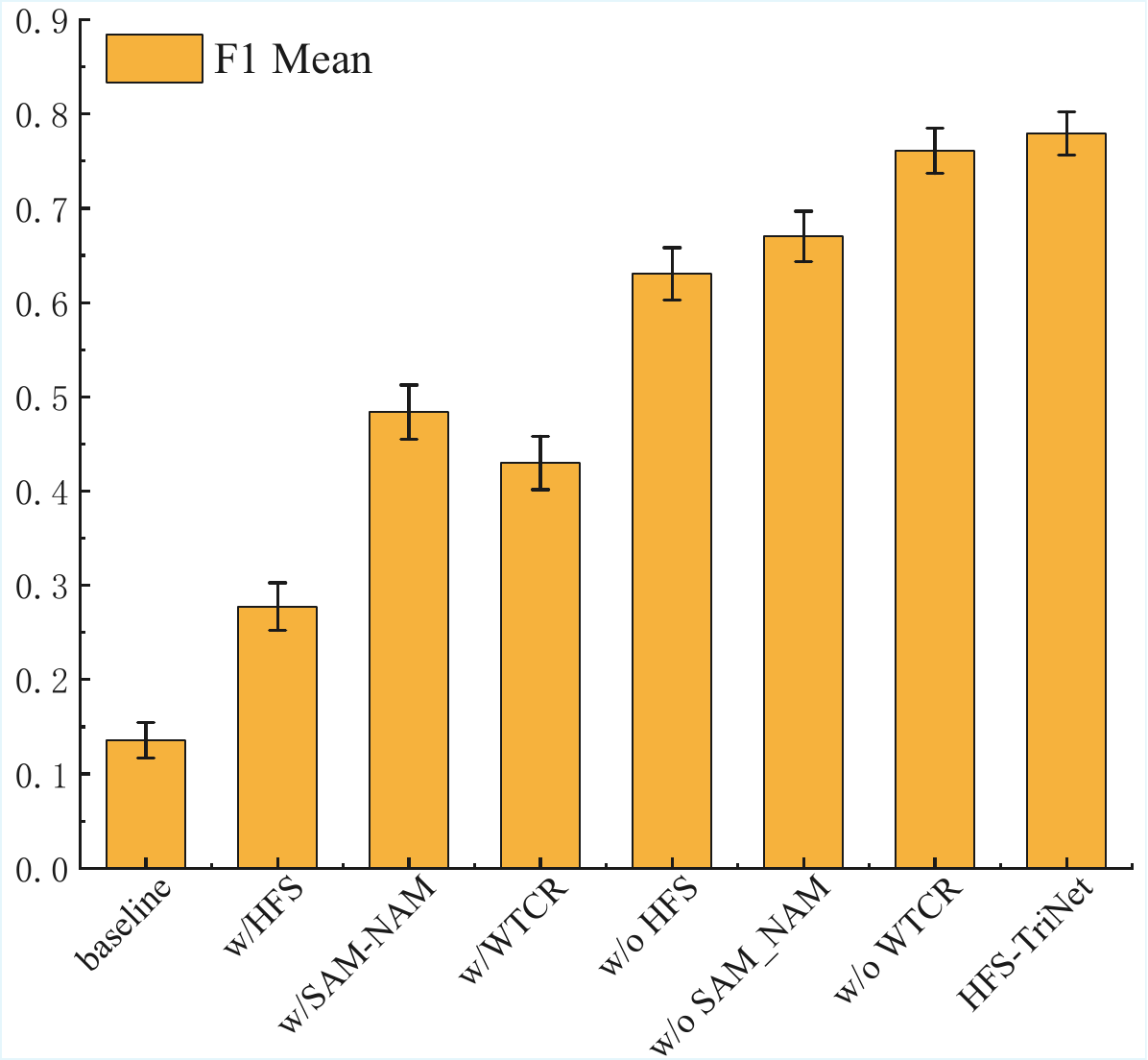}
	}
	\subfigure[]{
		\includegraphics[width=0.46\linewidth]{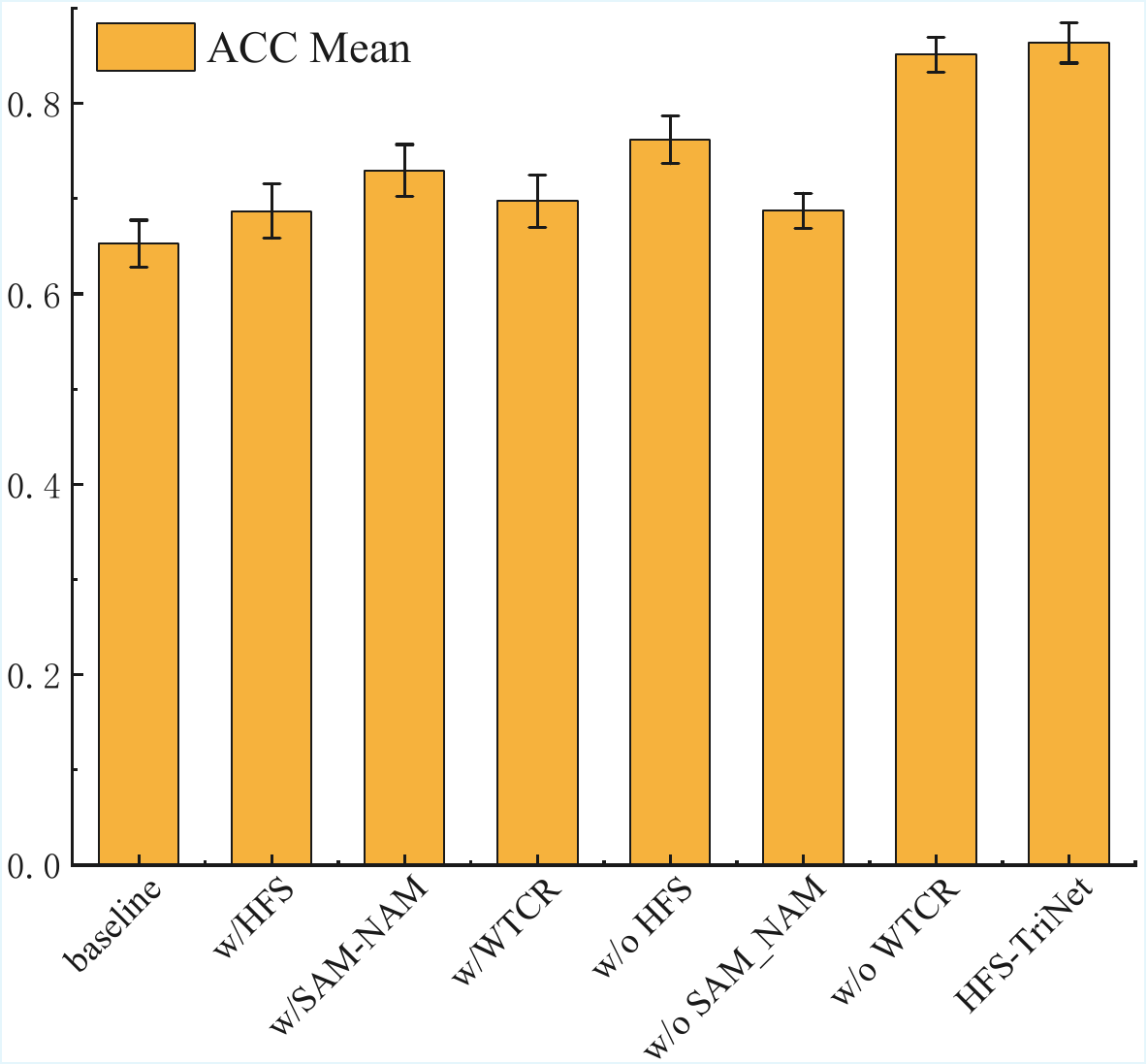}
	}
	\caption{\blue{
Bar plots of mean F1-score and accuracy (ACC) with corresponding $95\%$ confidence intervals for the full HFS-TriNet model and its ablated variants.}
}
	\label{fig:error-bar}
\end{figure}

This ablation study systematically evaluates the functional roles and synergistic contributions of the three core components of HFS-TriNet--—HFS, SAM\_NAM, and WTCR--—through different component combinations.
The results are summarized in Table~\ref{tab:modules}, where all metrics are reported as mean $\pm$ standard deviation of confidence intervals estimated via bootstrap resampling.

\par
Compared with the baseline model ${M\textsubscript{0}}$, introducing the HFS strategy alone (${M\textsubscript{1}}$) leads to clear improvements in F1-score and sensitivity, indicating that HFS effectively mitigates temporal redundancy and increases the likelihood of sampling lesion-relevant frames.
Incorporating SAM\_NAM alone (${M\textsubscript{2}}$) yields substantial gains in F1-score and AUC, highlighting the benefit of semantic feature adaptation under high intra-class similarity.
Similarly, the WTCR-only variant (${M\textsubscript{3}}$) improves both F1-score and AUC, demonstrating the value of frequency-aware modeling for speckle-dominated TRUS videos.

\par
Different module combinations exhibit distinct precision--sensitivity trade-offs.
For example, ${M\textsubscript{5}}$ (HFS + WTCR) favors sensitivity at the expense of precision, whereas ${M\textsubscript{6}}$ (SAM\_NAM + WTCR) shows the opposite tendency.
These results suggest that individual modules emphasize different aspects of the classification task and that relying on a single enhancement may bias the decision profile.

\par
By contrast, the full model ${M\textsubscript{7}}$ achieves a more balanced performance across accuracy, F1-score, AUC, precision, and sensitivity.
This indicates that the collaborative integration of HFS, SAM\_NAM, and WTCR enables complementary strengths to be jointly exploited while mitigating the limitations of individual components.

\par
Fig.~\ref{fig:error-bar} further illustrates these trends using mean performance with corresponding $95\%$ confidence intervals for accuracy and F1-score.
The reduced variability observed for the full model highlights improved statistical stability in addition to higher average performance.

\subsubsection{Effectiveness and Analysis of HFS}


\begin{figure}[htbp]
	\centering
	\includegraphics[width=\linewidth]{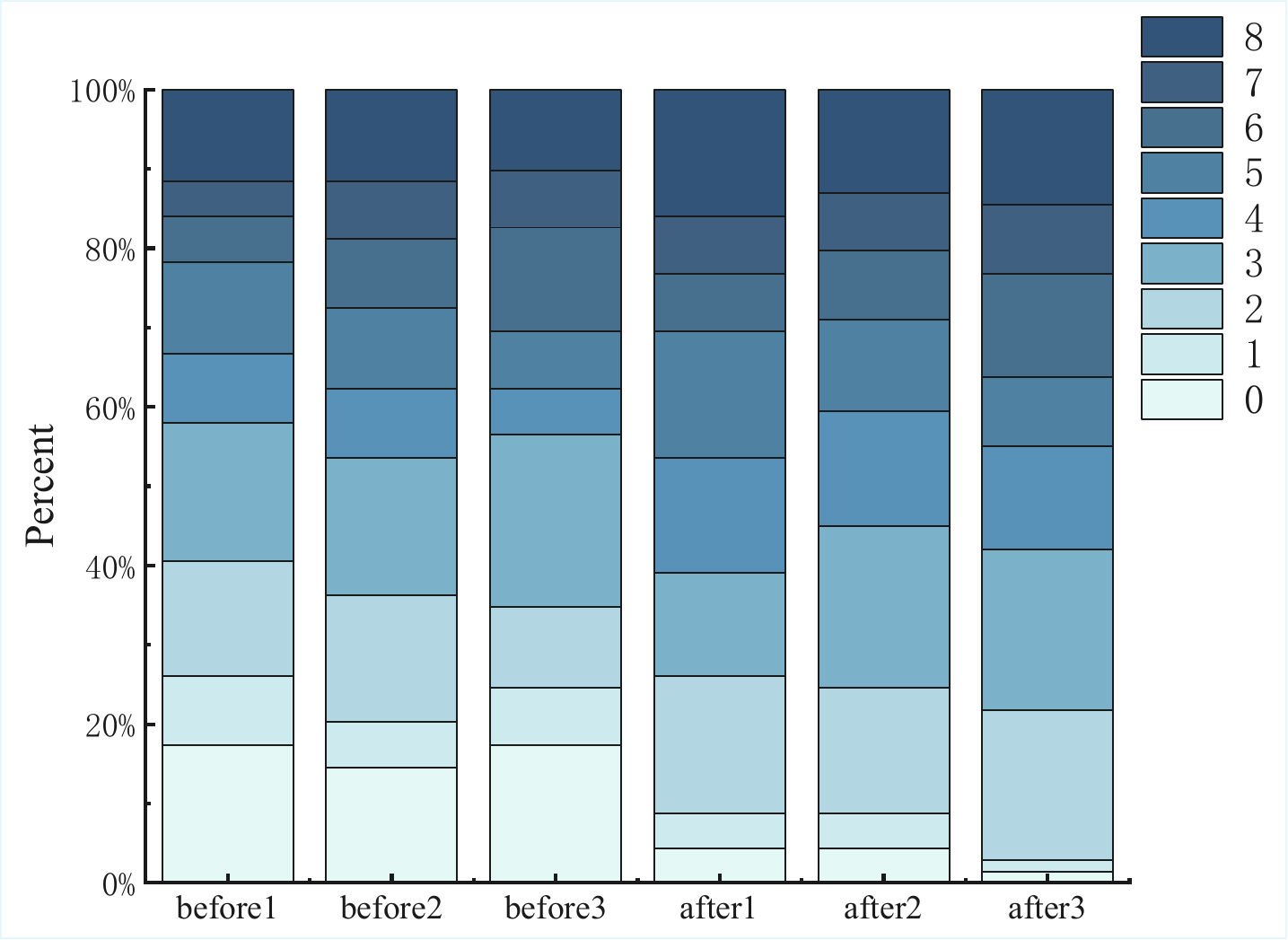}
	\caption{
Distribution of lesion-relevant frames per clip before and after applying the proposed HFS strategy.
Each bar represents one sampling trial, with colors indicating the number of annotated lesion-related frames per clip (0--8).
Compared with random sampling (\textit{before1--3}), HFS (\textit{after1--3}) shifts the distribution toward clips containing more lesion-relevant frames and reduces background-only clips.
}
	\label{fig:HFS}
\end{figure}

To quantitatively evaluate the effectiveness of the proposed HFS strategy, we conducted a controlled sampling experiment on a subset of 69 malignant non-diffuse TRUS videos.
In these videos, lesion-related information appears sparsely and intermittently, and a limited number of frames were manually annotated as lesion-relevant frames.
This subset was used exclusively for analyzing the sampling behavior of HFS, rather than for model training or performance evaluation.

For each video, multiple clips were generated by sampling frames according to Eq.~(\ref{eq:hfs}).
We compared two sampling strategies: random initialization of clip starting points (\textit{before}) and the proposed heuristic frame sampling (\textit{after}).
To account for randomness, the sampling process was repeated 20 times for each strategy.
For each sampled clip, we counted the number of annotated lesion-relevant frames and analyzed their distribution across all clips.

Quantitative results show that, after applying HFS, the proportion of clips containing more than three lesion-relevant frames increased markedly from 58.6\% to 79.3\%.
At the same time, the proportion of clips containing only background frames decreased substantially from 26.1\% to 2.9\%.
These results indicate that HFS significantly improves the efficiency of sampling lesion-relevant content under fixed clip-length constraints.

For intuitive illustration, Fig.~\ref{fig:HFS} visualizes three representative sampling trials before and after applying HFS.
Each vertical bar corresponds to one trial, and the horizontal segments indicate the distribution of lesion-relevant frames per clip.
Compared to random sampling, HFS produces a clear shift toward clips with richer lesion-related content and reduced variability across trials.
This qualitative trend is consistent with the quantitative analysis and confirms that HFS provides more stable and informative clip sampling for downstream TRUS video classification.

\subsubsection{Ablation Study on SAM Feature Fusion Position}
\begin{figure}[htbp]
	\centering
	\includegraphics[width=\linewidth]{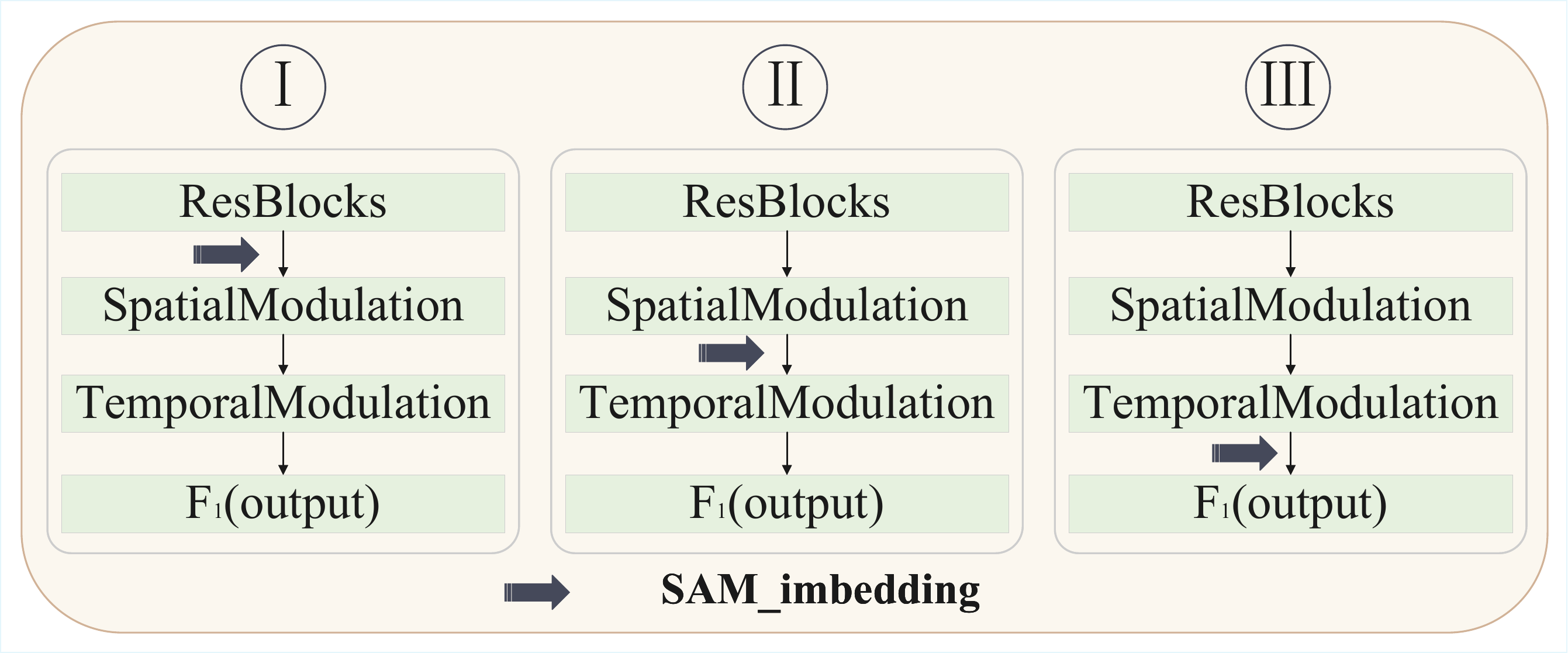}
	\caption{
Illustration of three fusion strategies for integrating SAM embeddings into the TPN backbone: fusion after the residual blocks (Strategy~\Rmnum{1}), after spatial modulation (Strategy~\Rmnum{2}), and after temporal modulation (Strategy~\Rmnum{3}).
}

	\label{fig:SAMfusion}
\end{figure}

\begin{table*}[htbp]
	\centering
	\caption{Ablation experiments on different SAM feature fusion strategies.
	 }
	\label{tab:SAMfusion}
	\setlength{\tabcolsep}{9pt}
	\begin{tabular}{ccccccc}
		\Xhline{1pt}
		\textbf{Method} & \textbf{Acc↑} & \textbf{F1-score↑} & \textbf{AUC↑} & \textbf{Precision↑} & \textbf{Sensitivity↑} & \textbf{Specificity↑} \\
		\Xhline{0.8pt}
		\uppercase\expandafter{\romannumeral1} & $0.6814 \pm 0.0257$ & $0.2285 \pm 0.0412$ & $0.5050 \pm 0.0354$ & \pmb{$0.9485 \pm 0.0165$} & $0.1301 \pm 0.0304$ & \pmb{$0.9954 \pm 0.0506$} \\
		\uppercase\expandafter{\romannumeral2} & $0.7046 \pm 0.0315$ & $0.4519 \pm 0.0257$ & $0.7231 \pm 0.0208$ & $0.6857 \pm 0.0207$ & $0.3354 \pm 0.0211$  & $0.9142 \pm 0.0205$ \\
		\uppercase\expandafter{\romannumeral3} & $0.7315 \pm 0.0114$ & $0.0211 \pm 0.0201$ & $0.7653 \pm 0.0182$ & $0.7374 \pm 0.0064$ & \underline{$0.4038 \pm 0.0216$} & $0.9182 \pm 0.0290$ \\
		\Xhline{0.8pt}
		\pmb{HFS-TriNet} & \pmb{$0.8622 \pm 0.0289$} & \pmb{$0.7766 \pm 0.0467$} & \pmb{$0.8226 \pm 0.0297$} & \underline{$0.9379 \pm 0.0288$} & \pmb{$0.6592 \pm 0.0714$} & \underline{$0.9720 \pm 0.0158$} \\
		\Xhline{1pt}
	\end{tabular}
\end{table*}

\begin{figure}[htbp]
	\centering
	\includegraphics[width=\linewidth]{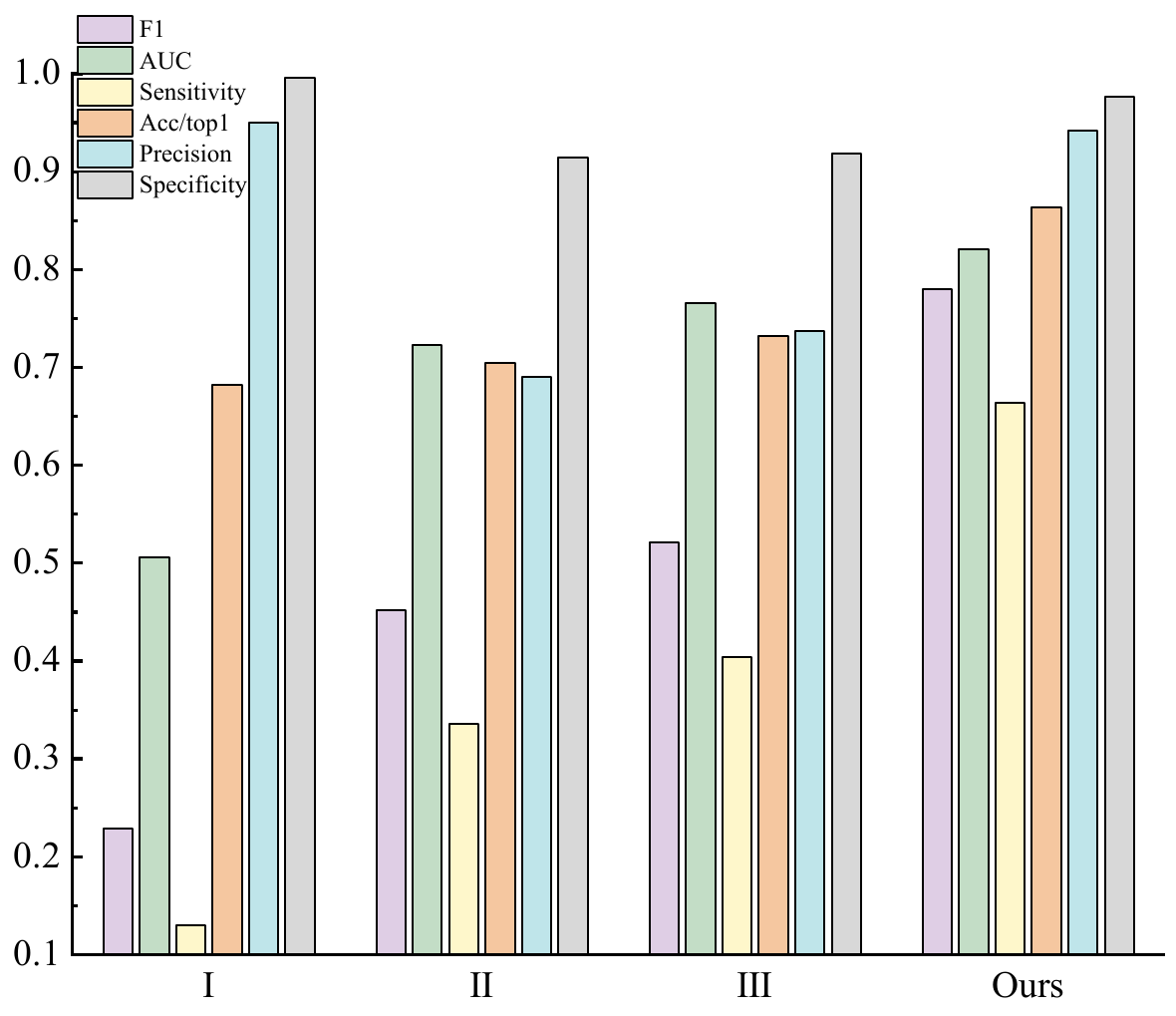}
	\caption{
\blue{Bar chart comparing the performance of different SAM feature fusion strategies across multiple evaluation metrics.
HFS-TriNet exhibits a more balanced performance profile than alternative fusion strategies.}
}
	\label{fig:SAMresult}
\end{figure}

To investigate how the fusion position of SAM-derived semantic features influences TRUS video classification, we evaluate three fusion strategies within the TPN backbone: fusion after the residual blocks (Strategy~\Rmnum{1}), after spatial modulation (Strategy~\Rmnum{2}), and after temporal modulation (Strategy~\Rmnum{3}), as illustrated in Fig.~\ref{fig:SAMfusion}.

\par
The quantitative results are summarized in Table~\ref{tab:SAMfusion}, where all metrics are reported as mean $\pm$ standard deviation estimated via bootstrap resampling.
Strategy~\Rmnum{1} exhibits high precision and specificity but extremely low sensitivity, resulting in poor F1-score, indicating that early semantic fusion limits effective malignant case identification.
Strategy~\Rmnum{2} substantially improves sensitivity and F1-score, suggesting that fusing semantic features after spatial modulation provides more effective guidance, though the precision--sensitivity balance remains suboptimal.

\par
Strategy~\Rmnum{3}, which introduces SAM features after full spatio-temporal modulation, achieves further improvements in F1-score and AUC while maintaining a more balanced precision--sensitivity trade-off.
These results indicate that delayed semantic fusion better complements temporal modeling in TRUS videos.

\par
Building upon Strategy~\Rmnum{3}, the final HFS-TriNet additionally incorporates normalization-based attention and the WTCR branch, achieving the best overall performance across all metrics.
As shown in Fig.~\ref{fig:SAMresult}, HFS-TriNet demonstrates more uniform performance across evaluation criteria, reflecting improved balance and stability.

\label{sec:experiments}

\section{Discussion}
\subsection{Clinical Relevance and Practical Considerations}
\blue{The proposed HFS-TriNet is designed as a computer-aided decision support framework for TRUS video analysis, rather than a standalone diagnostic system.
Its goal is to assist clinicians in video-level prostate cancer risk assessment under challenging TRUS imaging conditions, including severe speckle noise, weak lesion contrast, and high temporal redundancy.}

\blue{Under a fixed decision threshold, the proposed method achieves a sensitivity of approximately 66\%, indicating that some malignant cases may remain undetected.
This reflects the intrinsic difficulty of TRUS-based prostate cancer classification, particularly for small or diffuse lesions.
At the same time, the achieved AUC suggests that HFS-TriNet supports flexible operating-point adjustment.
In practice, the decision threshold may be tuned to prioritize sensitivity or specificity depending on the intended clinical scenario, which is application-dependent and beyond the scope of the present study.}

\blue{From a computational perspective, HFS-TriNet is intended for offline or near–real-time analysis rather than frame-by-frame real-time deployment.
With an inference throughput of approximately 14--15 videos per second under batch processing, the proposed framework is compatible with typical post-acquisition TRUS workflows, such as secondary review or biopsy planning, without requiring changes to existing acquisition protocols.}

\blue{The robustness of the proposed method is supported by multiple complementary evaluation strategies.
Strict patient-level partitioning was enforced to prevent data leakage, and performance was assessed using both bootstrap-based confidence interval estimation and patient-level five-fold cross-validation.
The consistent results observed across different patient splits indicate stable performance within the studied multi-institutional cohort.}

\blue{Importantly, the contribution of this work lies in a problem-driven, system-level integration of complementary representations.
Heuristic frame sampling addresses temporal redundancy and variable video length, the semantic branch introduces high-level anatomical priors, and the frequency-aware branch enhances robustness to speckle noise.
Ablation studies demonstrate that their coordinated interaction yields more balanced performance than any single component alone.}

\subsection{Limitations and Future Work}
\blue{Several limitations should be acknowledged.
First, the dataset was retrospectively collected from a limited number of institutions, and broader multi-center validation is required to further assess generalizability.
Second, the current study relies on video-level labels without fine-grained lesion annotations or pathological grading, which limits clinically stratified analysis.
Third, although competitive sensitivity is achieved, further improvements may be possible through cost-sensitive learning, alternative loss formulations, or threshold calibration.
In addition, failure cases were observed in TRUS videos with extremely subtle lesion appearance or severe speckle noise, where discriminative cues remain limited even at the video level, reflecting inherent challenges of ultrasound imaging rather than model-specific deficiencies.
Finally, the robustness of foundation-model features under controlled speckle perturbations was not explicitly evaluated and warrants further investigation.}
\label{sec:discuss}

\section{Conclusion}

In this study, we present HFS-TriNet, a problem-driven and system-level framework for prostate cancer classification from TRUS videos.
By combining a heuristic frame selection strategy with collaborative spatio-temporal, semantic, and frequency-aware feature learning, the proposed method addresses key challenges inherent to TRUS video analysis, including temporal redundancy, low signal-to-noise ratio, and high intra- and inter-class similarity.
Extensive experiments on a multi-institutional TRUS video dataset, including patient-level evaluation, ablation studies, statistical significance analysis, and cross-validation, demonstrate that HFS-TriNet achieves strong and balanced performance across multiple evaluation metrics.
Rather than introducing new learning operators, this work emphasizes a clinically motivated, system-level integration of complementary representations tailored to TRUS video characteristics.
We believe that the proposed framework provides a practical and reproducible step toward computer-aided TRUS video analysis and offers useful insights for future research on ultrasound-based video understanding.
\blue{The source code of HFS-TriNet will be publicly released upon acceptance to facilitate reproducibility and further research.}

\label{sec:conclusion}

\section{Ethical Spproval}
This work involved human subjects in its research. Approval of all ethical and experimental procedures and protocols was granted by the Shenzhen People's Hospital Medical Ethics Committee under Application No. LL-KY-2025131-01, and was conducted in accordance with the ethical principles outlined in the "Administrative Measures for Ethical Review of Biomedical Research Involving Humans" (2016) issued by the Ministry of Health of China, the World Medical Association's Declaration of Helsinki, and the CIOMS International Ethical Guidelines for Biomedical Research Involving Human Subjects. This retrospective study was approved by the institutional review board with waiver of informed consent.

\section*{References}
\bibliographystyle{IEEEtran}
\bibliography{references}

\end{document}